\documentclass[conference]{IEEEtran}
\IEEEoverridecommandlockouts

\usepackage{amsmath,amssymb,amsfonts}
\usepackage{algorithmic}
\usepackage{graphicx}
\usepackage{textcomp}
\usepackage{xcolor}
\usepackage{url}
\usepackage{tabularx}
\usepackage{bm}
\usepackage{upgreek}
\usepackage{booktabs}
\usepackage{array}
\usepackage{multirow}
\usepackage{graphicx}
\usepackage{svg}
\usepackage{wasysym}
\usepackage[flushleft]{threeparttable}
\usepackage{enumitem}
\usepackage{cite}
\usepackage{hyperref}

\def\secref#1{Section~\ref{#1}}
\def\figref#1{Fig.~\ref{#1}}
\def\tabref#1{Table~\ref{#1}}
\def\eqref#1{(\ref{#1})}

\def\where{\noindent where }

\newcolumntype{A}{>{\centering\arraybackslash}X}

\def\BibTeX{{\rm B\kern-.05em{\sc i\kern-.025em b}\kern-.08em T\kern-.1667em\lower.7ex\hbox{E}\kern-.125emX}}

\makeatletter
\newcommand{\newlineauthors}{%
  \end{@IEEEauthorhalign}
  \hfill\mbox{}\par
  \mbox{}\hfill\begin{@IEEEauthorhalign}
}
\makeatother

\title{KGIF: Optimizing Relation-Aware Recommendations with Knowledge Graph Information Fusion}

\author{
    \IEEEauthorblockN{
        Dong Hyun Jeon\textsuperscript{1}, 
        Wenbo Sun\textsuperscript{2},  
        Houbing Herbert Song\textsuperscript{3},
        Dongfang Liu\textsuperscript{4}, \\
        Velasquez Alvaro\textsuperscript{5}, 
        Yixin Chloe Xie\textsuperscript{6}, 
        Shuteng Niu\textsuperscript{1}
    }

    \IEEEauthorblockA{
        \textit{\textsuperscript{1} Department of Computer Science, Bowling Green State University}\\
        \textit{\textsuperscript{2} Transportation Research Institute, University of Michigan - Ann Arbor}\\
        \textit{\textsuperscript{3} Department of Information Systems. University of Maryland, Baltimore County}\\
        \textit{\textsuperscript{4} Department of Computer Engineering, Rochester Institute of Technology}\\
        \textit{\textsuperscript{5} College of Engineering \& Applied Science, University of Colorado Boulder}\\
        \textit{\textsuperscript{6} Department of Information Technology, Kennesaw State University}
    }
}

\begin{document}

\maketitle

\begin{abstract}

While deep-learning-enabled recommender systems demonstrate strong performance benchmarks, many struggle to adapt effectively in real-world environments due to limited use of user-item relationship data and insufficient transparency in recommendation generation. Traditional collaborative filtering approaches fail to integrate multifaceted item attributes, and although Factorization Machines account for item-specific details, they overlook broader relational patterns. Collaborative knowledge graph-based models have progressed by embedding user-item interactions with item-attribute relationships, offering a holistic perspective on interconnected entities. However, these models frequently aggregate attribute and interaction data in an implicit manner, leaving valuable relational nuances underutilized.

This study introduces the Knowledge Graph Attention Network with Information Fusion (KGIF), a specialized framework designed to merge entity and relation embeddings explicitly through a tailored self-attention mechanism. The KGIF framework integrates reparameterization via dynamic projection vectors, enabling embeddings to adaptively represent intricate relationships within knowledge graphs. This explicit fusion enhances the interplay between user-item interactions and item-attribute relationships, providing a nuanced balance between user-centric and item-centric representations. An attentive propagation mechanism further optimizes knowledge graph embeddings, capturing multi-layered interaction patterns. The contributions of this work include an innovative method for explicit information fusion, improved robustness for sparse knowledge graphs, and the ability to generate explainable recommendations through interpretable path visualization. The implementation and datasets for this study are publicly available\footnote{\url{https://github.com/ryandhjeon/KGIF}}.

\end{abstract}

\begin{IEEEkeywords}
Recommender systems, Embedding Fusion, Collaborative Filtering, Knowledge Graph, Graph Attention Network
\end{IEEEkeywords}

\section{Introduction}

Recommender systems have evolved beyond simple business tools; they now form the backbone of personalized digital experiences, fundamentally shaping how users interact with content, products, and services. These systems have the power to curate experiences that feel tailored to individual preferences, enhancing user satisfaction and engagement. With the surge of advanced deep learning models, recommender systems have gained the ability to process vast, complex datasets, delivering highly relevant and personalized recommendations~\cite{10.1145/3639063}. However, despite significant advances, real-world challenges such as sparse data and the need for interpretable recommendations continue to limit their effectiveness. Traditional collaborative filtering (CF) methods~\cite{resnick1994grouplens}, while historically successful, often fail to fully utilize the rich, auxiliary information embedded in item attributes and user behavior, resulting in suboptimal performance when data is sparse.

Recent developments in recommendation technologies have focused on integrating side information—beyond mere item IDs—such as item attributes and user interactions, leading to more refined and effective models~\cite{bordes2013translating, wang2014knowledge, lin2015learning}. Collaborative knowledge graphs (CKGs) have emerged as a powerful framework to capture these complex relationships, especially in scenarios where user-item interactions are sparse and item-attribute interactions are intricate~\cite{yang2022knowledge, yang2023knowledge, wang2019kgat}. However, while CKGs have advanced the representation of recommendation data, significant challenges remain in learning robust embeddings in the latent semantic space, primarily due to the heterogeneous nature of nodes and relations within the graph.

Graph Neural Networks (GNN)-based recommender systems have shown great promise by constructing high-quality embeddings for diverse data types~\cite{gao2022graph}, capturing high-order connectivity for intricate reasoning processes~\cite{li2021higher}, and encoding semi-supervised signals to combat sparse supervision~\cite{dai2022towards}. However, these methods often underperform when it comes to capturing distant neighbors crucial for multi-hop reasoning, limiting their ability to fully exploit relational data. To address these issues, Graph Attention Networks (GAT)-based models have pushed the envelope further. For example, models like Collaborative Knowledge-aware Attentive Network (CKAN)\cite{wang2020ckan} and Knowledge Graph-based Intent Network (KGIN)\cite{wang2021learning} have excelled at leveraging attention mechanisms to weigh the importance of neighboring nodes, thus improving recommendation accuracy. Knowledge-enhanced Graph Contrastive Learning (KGCL)~\cite{yang2022knowledge} goes further by introducing knowledge graph augmentation, which reduces noise in information aggregation, resulting in more robust item representations. However, these models still suffer from two critical limitations: (1) they often treat users and entities in the CKG as homogeneous nodes, neglecting the multifaceted nature of user-item interactions, and (2) they implicitly incorporate item-attribute information during message passing, potentially overlooking crucial relational insights.

To address these challenges, we propose the \textbf{K}nowledge \textbf{G}raph Attention Network with \textbf{I}nformation \textbf{F}usion (KGIF), a robust framework designed to explicitly fuse both entity and relation information, leveraging a self-attention mechanism to enhance recommendation quality. In KGIF, we introduce a reparameterization technique through dynamic projection vectors, where each relation in the knowledge graph is represented by distinct head and tail projection vectors. This enables our model to capture the complex, heterogeneous relationships within the graph more effectively, ensuring that both entity and relational embeddings are fully utilized in the recommendation process. The fusion of entity and relation embeddings is further refined by an attentive embedding propagation layer, which allows the model to capture high-order connectivity, thus improving the ability to handle complex and sparse data.

The effectiveness of our approach is demonstrated through extensive experiments on benchmark datasets such as Amazon-book~\cite{he2016ups}, Last-FM~\cite{Celma:Springer2010
}, and Yelp2018~\cite{wang2019neural}, where KGIF consistently outperforms existing state-of-the-art methods. Our main contributions to this paper are as follows:

\begin{itemize}
    \item We introduce a robust information fusion method, KGIF, using dynamic projection vectors for heterogeneous relations.
    \item Outperforming the state-of-the-art (SOTA) by a substantial margin, as evidenced by extensive comparative experiments across three benchmarks.
    \item A visual interpretation of the recommendation-making process, enhancing the transparency of decision-making.
\end{itemize}

The rest of this paper is organized as follows: \secref{RelatedWork} provides an overview of relevant research; \secref{Methodology} details the proposed KGIF framework and its technical specifications; \secref{Experiments} outlines the experimental setup and results, with a detailed comparative analysis; and \secref{Conclusion} summarizes the main findings and suggests directions for future work.


\begin{figure}[t]
    \centering \includegraphics[width=1\columnwidth]{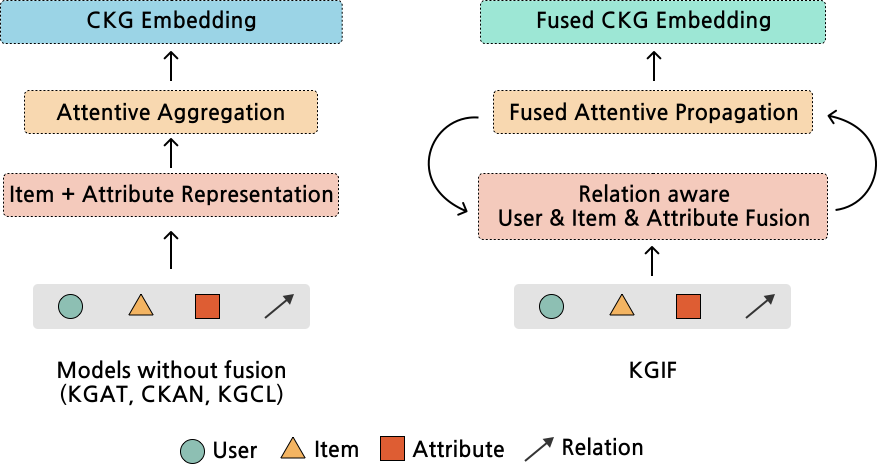}
    \caption{\small{The SOTA recommender models work without fusion. KGIF utilizes not only attribute but also relation information to update both user and item representation for fused CKG embedding.}}
    \label{fig:Core}
\end{figure}

\section{Related Work}\label{RelatedWork}

\begin{figure*}[t]
    \centering
    \includegraphics[width=0.9\textwidth]{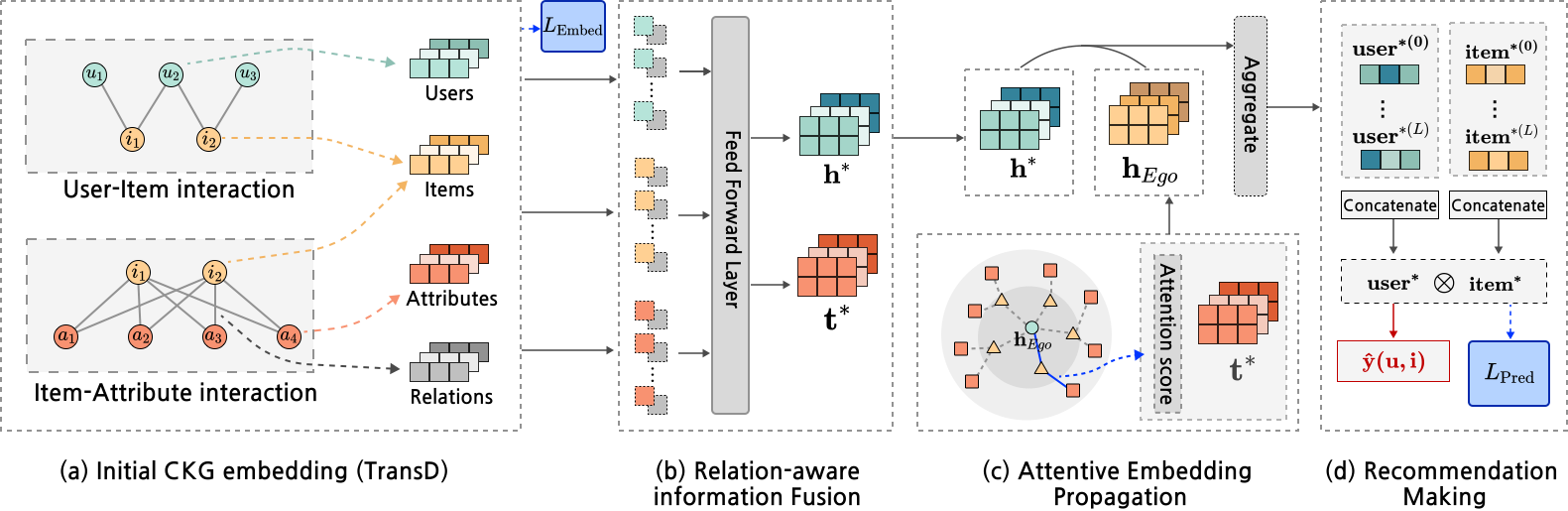}
    \caption{\small{Overall process of KGIF. (a) Learn a set of initial CKG embedding with TransD. (b) Information fusion by using projection vectors from the previous step. (c) Attentive embedding propagation with self-attention of triplets for updating CKG embedding. (d) Recommendation-making based on user-item information aggregation.}}
    \label{fig:Overview}
\end{figure*}

\subsection{CKG based Recommendation}

CKG-based recommendations aim to leverage knowledge graphs to items to the users with their attributes, and recent advancements~\cite{bordes2013translating, wang2019kgat, lin2015learning} include a significant focus on integrating side information. Unlike previous approaches that primarily utilized item IDs as the sole attribute, these newer methods tried to better represent recommendation data that has sparse user-item interactions and complex item-attribute interactions. Intuitively, incorporating such closely related information could enhance recommendations. 

As shown in \figref{fig:ckg}, CKGs contain two structures: 1) a user-item bipartite graph (BG) and 2) a knowledge graph that captures complex relations between items and attributes. Within CKGs, GAT-based models~\cite{velivckovic2017graph} have made promising progress with sparse yet highly complex data. As an example, KGAT~\cite{wang2014knowledge} provided an end-to-end model to explore high-order linear and non-linear relations in the data with a self-attention mechanism. KGAT performs propagation over CKG to complement entity embeddings with an attention mechanism. Importantly, it offered a visual interpretation of the recommendation-making process and utilization of side information. However, their method treats user, item, and attribute embeddings as the same entity nodes in CKG, which reduces the heterogeneity of the knowledge graph. Since the types of users and items may be extremely broad, the high-order interactions may contain high noise within the latent embeddings. Similar approaches like CKAN~\cite{wang2020ckan} and KGCL~\cite{zhang2023kgcl}, each with their unique methodologies, made decent improvements on recommendations with sparse data. 

CKG-based methods have demonstrated their superiority and become the primary focus in recent recommender systems. However, these methods share several common problems: 1) unable to distinguish users and entities in CKGs, 2) relying on the homogeneity assumption between User-Item Interaction BG and KG, leading to potential noise in embeddings, and 3) do not guarantee the proper utilization of side information with implicit integration in message passing. These challenges can lead to the ineffective use of side information and inaccurate modeling of user-item interactions.  

\subsection{Information Fusion for Recommendation}

As CKG-based recommendation models have extensively explored side information-aware recommendations, there has also been significant exploration into side information fusion with self-attention blocks. This involves integrating side information embeddings with item embeddings prior to the attention layer, enabling the attentive learning process to consider side information in the final representation. Fusing side information for recommendation systems~\cite{zhang2019feature, zhou2020s3} not only significantly improves model performance by allowing models to gain a more accurate and comprehensive understanding of user-item relationships, but also can beneficially help the next-item prediction. Such additional information is pivotal in refining KG embeddings and addressing traditional data sparsity issues in deep learning-enabled models.

A classic model, Parallel recurrent neural network (p-RNN)~\cite{hidasi2016parallel}, simply concatenated the side information into the item entity representations, and ICAI-SR~\cite{yuan2021icai} uses the idea of neighborhood aggregation from GNN by aggregating the item and attribute representations from the complex relations between items and attributes. Rather than fusing the side information in the input layer, DIF-SR~\cite{xie2022decoupled} separates the attention calculation process for different types of side information, creating fused attention matrices. However, these modules cannot address the main limitations of recent works on CKG-based recommendations.

\section{Methodology} \label{Methodology}

As demonstrated in \figref{fig:Overview}, KGIF consists of four hierarchically connected components: a) initial Embedding, where the data is reformulated as a CKG, and initial embeddings are generated for entities and relations b) relation information fusion, where relation-specific information is explicitly fused with entity embeddings to enrich the representations for downstream tasks c) attentive embedding propagation, where the fused information is passed through an attention mechanism to enhance the propagation of relevant information across the graph and d) recommendation making, where the processed embeddings are used to generate personalized recommendations based on user-item interactions. This section delves into the technical details of each component.

\subsection{Problem Formulation}

\begin{figure}[t]
    \centering
    \includegraphics[width=0.9\columnwidth]{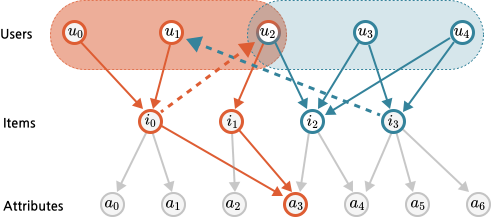}
    \caption{\small{Comparison between CF and CKG. CF recommends items solely based on user-item interactions, while CKG incorporates item attributes in addition to user-item interactions when making recommendations to users.}}
    \label{fig:ckg} 
\end{figure}

Primarily, we formulate the data as a CKG ($\mathcal{G}_\text{CKG}$) that has two graph structures, a user-item bipartite graph ($\mathcal{G}_\text{UI}$) and an item-entity knowledge graph ($\mathcal{G}_\text{KG}$). As shown in \figref{fig:ckg}, $\mathcal{G}_\text{UI}$ represents direct user-item interactions, commonly used in CF methods, which identify similar users based on shared items. For example, users $u_2$, $u_3$, and $u_4$ all share item $i_2$, making them similar, so item $i_3$ is recommended to user $u_1$. However, this approach becomes less effective with sparse data, where users like $u_1$ and $u_2$ do not share common items. In a CKG, $\mathcal{G}_\text{KG}$ provides item properties encoded in entities, allowing for the identification of similar users based on indirect connections. For instance, users $u_0$, $u_1$, and $u_2$ are identified as similar based on a shared property $a_3$ from items $i_0$ and $i_1$, leading to the recommendation of item $i_0$ to user $u_2$.

BG is a special graph structure where nodes can be divided into two disjoint sets such that no two nodes within the same set are adjacent. In this work, we define $\mathcal{G}_\text{UI}$ as a bipartite graph  with a single relation: $\mathcal{G}_\text{UI} = \{(u, y_{ui},i|u \in \mathcal{U}, i \in \mathcal{I}, y_{ui} \in \mathcal{Y}\},$ where $\mathcal{U}$ denotes the user set containing $M$ users, $\mathcal{I}$ denotes the item set with $N$ items, and $\mathcal{Y} \in \{0,1\}$ is a user-item interaction set with:

\vspace{-5mm}
\begin{equation} 
    y_{ui} = 
    \begin{cases}
        1, & \text{if an interaction exists between a user and an item}\\
        0, & \text{otherwise}
    \end{cases}.
\end{equation}

Moreover, the data also contains rich side information on items, including but not limited to titles, authors, and publishers. Importantly, the item attributes do not have direct connections to the users. In this study, we represent these information with $\mathcal{G}_\text{KG}$ defined as: $\mathcal{G}_\text{KG} = \{(h,r,t|h,t \in \mathcal{I} \cup \mathcal{A}, r \in \mathcal{R}\}$, where $\mathcal{A}$ represents item attribute set with $K$ attributes and $\mathcal{R}$ denotes relation set with $L$ relation types. Furthermore, we combine $\mathcal{G}_\text{UI}$ and $\mathcal{G}_\text{KG}$ to define $\mathcal{G}_\text{CKG}$ as: $\mathcal{G}_{\text{CKG}} = \{(h,r,t)|h,t \in \mathcal{E}, r \in \mathcal{R'}\}$, where $\mathcal{E} = \mathcal{U} \cup \mathcal{I} \cup \mathcal{A}$ and $\mathcal{R'} = \mathcal{Y} \cup \mathcal{R}$.

Finally, we formulate the objective of this work as KG completion for the user-item BG by using the side information provided by the item-entity KG.

\subsection{CKG Embedding with TransD}

The raw CKG must first be converted into a latent semantic space to facilitate easier manipulation. Fusing the side information with relationships at the beginning can introduce unnecessary complexity and computational overhead. Therefore, we adopt TransD~\cite{ji2015knowledge}, a simple yet effective translational embedding method, for the initial CKG embedding. TransD is particularly suited to multi-relational and sparse knowledge graphs because it dynamically adjusts entity embeddings based on relational contexts. This is a critical improvement over methods like TransR~\cite{lin2015learning}, which rely on static projection matrices that fail to adapt to the diverse patterns of relations within the graph. In TransR, the same projection matrix is applied to all entity embeddings, regardless of the relation, resulting in a generalized transformation that overlooks relation-specific characteristics. Consequently, the model struggles to differentiate between distinct relational contexts.

This issue is addressed by introducing \textbf{dynamic projection vectors} for both head (users or items) and tail (items or attributes) entities. These vectors allow the embeddings to be adjusted depending on the specific relation connecting them. The projection matrices are constructed using projection vectors $\mathbf{h}_p$, $\mathbf{r}_p$, and $\mathbf{t}_p$, which dynamically transform the embeddings of both the head and tail entities based on the relation, ensuring that each interaction is contextually tailored. This dynamic adjustment ensures that the entity embeddings reflect the intricacies of varying item-attribute interactions, which is critical for accurately capturing relational complexity.

However, while this method effectively preserves the structure of a CKG, it primarily focuses on direct relationships between users and items. It lacks the capability to emphasize indirect connections—such as shared attributes between items (e.g., genre, author)—and does not dynamically adjust the embeddings for multi-hop reasoning, which can reveal latent user preferences. This limitation is addressed in the following section, which enhances the fusion of entity and relation embeddings during the reparameterization process.

TransD represents each triplet as $(\textbf{h}, \textbf{r}, \textbf{t})$, where $\mathbf{h}, \mathbf{t} \in \mathbb{R}^m$ are the learned embeddings of the head and tail entities, and $\mathbf{r} \in \mathbb{R}^n$ is the learned embedding of the relation. The dynamic projection vectors $\mathbf{h}_p$, $\mathbf{t}_p \in \mathbb{R}^m$ for the head and tail entities, and $\mathbf{r}_p \in \mathbb{R}^n$ for the relation, are used to construct projection matrices $\mathbf{M}^{h}_{r}$ and $\mathbf{M}^{t}_{r}$. These projection matrices allow the model to tailor the entity embeddings $\mathbf{h}$ and $\mathbf{t}$ to the specific relational context $r$, enabling the model to effectively capture intricate interactions between items and attributes across various types of relations within the CKG. The projection matrices for a relation $r$ are formulated as:

\begin{equation}
    \mathbf{M}^{h}_{r} = \mathbf{r}_p\mathbf{h}_p^\top + \mathbf{I}^{m \times n}, \quad
    \mathbf{M}^{t}_{r} = \mathbf{r}_p\mathbf{t}_p^\top + \mathbf{I}^{m \times n},
\end{equation}

\where $\mathbf{I}^{m \times n}$ is the identity matrix. The transformed embeddings $\mathbf{h}_{\perp}$ and $\mathbf{t}_{\perp}$ are calculated as:

\begin{equation}
    \mathbf{h}_{\perp} = \mathbf{M}^{h}_{r} \mathbf{h}, \quad \mathbf{t}_{\perp} = \mathbf{M}^{t}_{r} \mathbf{t}.
\end{equation}

Furthermore, given triplet (h,r,t), the possibility of evaluating the likelihood of a triplet’s existence, with the classic translation principle:

\begin{equation}
    g(h,r,t) = - \lVert \mathbf{h}_{\perp} + \mathbf{r} - \mathbf{t}_{\perp} \rVert^2_2.
\end{equation}

The embedding objective function $\mathcal{L}_{\text{Embed}}$ is formulated as:

\begin{equation}
    \mathcal{L}_{\text{Embed}}=\sum_{(h,r,t,t')\in{T}} - \ln \,\sigma\, \left( g(h,r,t')-g(h,r,t) \right),
\end{equation}

\where $\sigma(\cdot)$ is the sigmoid function, $(h,r,t)$ is the positive triplet, and $(h,r,t')$ is the negative triplet sampled from $\mathcal{T} = \{ (h,r,t,t'|(h,r,t) \in \mathcal{G_\text{CKG}}, \: (h,r,t') \notin \mathcal{G_\text{CKG}} \}$. The negative triplet, $(h,r,t')$ is constructed by replacing a tail in a positive triplet with a broken tail randomly. 

\subsection{Explicit Information Fusion Layer}

While the introduction of the initial CKG embedding provides a solid structural foundation, it does not fully leverage the complex, indirect connections between users and item attributes—such as genres or authors—that are critical for accurate recommendations. Li~\cite{li2021prefix} and Liu\cite{liu2021p} pointed out that reparameterized embeddings can suffer from performance degradation due to sensitivity to initialization. We address this issue by using dynamic projection vectors that adapt to relational contexts, providing greater stability during training. To overcome this limitation, we introduce an explicit information fusion method that integrates entity and relation embeddings using projection vectors from the initial embedding phase. This fusion aims to:

\begin{itemize}
    \item \textbf{Capture indirect connections:} Identify hidden links between users and items that are not immediately apparent through direct interactions but are essential for capturing latent preferences and generating more accurate predictions.
    \item \textbf{Contextualize user preferences:} Better understand how users' preferences are shaped by specific attributes, often overlooked in traditional embeddings, such as genre or author influences.
\end{itemize}

The fusion process updates the initial embeddings through the following transformation:

\begin{equation} 
    \label{eq:entityRelFusion}
    \mathbf{h}^{'} = \mathbf{h}_{\perp} \odot \mathbf{r}, \quad \mathbf{t}^{'} = \mathbf{t}_{\perp} \odot \mathbf{r},
\end{equation}

\where $\mathbf{h}^{'}$ and $\mathbf{t}^{'}$ are updated embeddings, and $\odot$ represents the Hadamard (element-wise) product. While this operation integrates entity-specific and relation-specific information, it is insufficient for capturing non-linear relationships. Therefore, we propose a non-linear reparameterization by passing the fused embeddings through learned transformation matrices and applying a ReLU activation function:

\begin{equation} 
    \mathbf{h}^{*} = \text{ReLU}(\mathbf{h}^{'} \mathbf{W}_1 + \mathbf{b}), \quad \mathbf{t}^{*} = \text{ReLU}(\mathbf{t}^{'} \mathbf{W}_2 + \mathbf{b}),
\end{equation}

\where $\mathbf{W}_{1}, \mathbf{W}_{2} \in \mathbb{R}^{n \times n}$ are the transformation matrices, and $\mathbf{b} \in \mathbb{R}^n$ is the bias vector. This final transformation allows the model to capture the non-linear interactions between entities and relations, leading to more accurate recommendations. Detailed analysis of this fusion method is presented in the ablation study (\secref{sec:embeddingFusion}).

\subsection{Attentive Embedding Propagation Layer}

Next, we develop a self-attention mechanism ~\cite{wang2019kgat,velickovic2017graph} with fused embeddings to accommodate high-order connectivity data by aggregating insights from the distant neighbors in the network. The self-attention aims to quantify the significance of each triplet by computing the attention scores for triplets, reflecting their connections' relative importance. The attention scores are used in the later recommendation-making layer to compute the probability of a user endorsing a specific item.

Contrary to Graph Convolutional Network (GCN)~\cite{kipf2016semi}, which utilizes a shared weight matrix for node-wise feature transformations, our model distinguishes neighbor nodes based on their importance by assigning different weights to individual nodes. This is achieved by a function that computes the self-attention score of each triplet. The function is expressed as:

\begin{equation} \label{eq:attentionWeight}
    \delta(h,r,t)=(\mathbf{M}^{t}_{r}\mathbf{t}^{*})^{\top}\tanh(\mathbf{M}^{h}_{r}\mathbf{h}^{*}+\mathbf{r}),
\end{equation}
\where $\mathbf{M}^{t}_{r}$ and $\mathbf{M}^{h}_{r}$ are the projection matrices from the initial CKG embedding layer, $\mathbf{h}^{*}$ and $\mathbf{t}^{*}$ are fused embeddings, $\tanh$ is the activation function. This attention score aims to capture the semantic distance between $h$ and $t$ in the relation $r$ space. Moreover, as shown in \figref{fig:ego}, we use the first-order ego-network, which is a subgraph focused on one node and its immediate connections, to model the relations in direct neighbors.  Then, normalized self-attention scores for all triplets in an ego network with the softmax function:

\begin{figure}[t] 
    \centering
    \includegraphics[width=0.4\columnwidth]{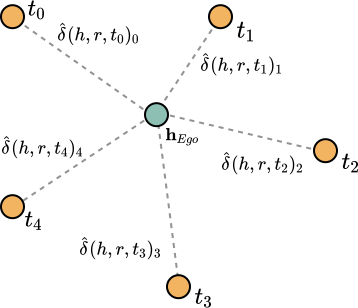}
    \caption{\small{First-order ego-network of a node $h$. The head entity is in the middle denoted as $\mathbf{h}_{Ego}$ and tails are denoted as $t_{0-4}$.}}
    \label{fig:ego}
\end{figure}

\begin{equation} 
    \label{eq:normalize}
    \hat{\delta}(h,r,t) = \frac{\text{exp}(\delta(h,r,t))}{\sum_{(h,r',t')\in{\mathbf{Ego}_{h}}}\text{exp}(\delta(h,r',t'))}.
\end{equation}

The normalized attention scores quantify the importance of surrounding neighbors of a node. The first-order characteristic of a node can be captured with a linear combination of $h$’s ego-network:

\begin{equation} 
    \label{eq:ego}
    \textbf{h}_{Ego} = \sum_{(h,r,t)\in{\mathbf{Ego}_{h}}} \hat{\delta}(h,r,t)\textbf{t}^{*}.
\end{equation}

Furthermore, the propagation layers can be stacked to explore a higher-order connectivity: 

\begin{equation} 
    \label{eq:highOrderConnectivity}
    \textbf{h}^{(l-1)}_{Ego} = \sum_{(h,r,t)\in{\mathbf{Ego}_{h}}} \hat{\delta}(h,r,t){\textbf{t}^{*}}^{(l-1)},
\end{equation}
\where the $\mathbf{t}^{*(l-1)}$ is the generated $t$ from the previous propagation layer and is memorized as the number of the layer $l$ increases. 


Finally, we aggregate the individual entity representations  $\mathbf{h}^*$, with the representations of their corresponding ego-networks, $\mathbf{h}_{Ego}$. This process results in the updated entity representation,  $\mathbf{h}^{(l)}=f(\mathbf{h}^{*}, \mathbf{h}_{Ego})$ representing the updated entity representation at the $l^{th}$ level of propagation layers. For the aggregation function, Bi-interaction Aggregator proposed by Wang \textit{et al}.~\cite{wang2019kgat} is identified as the most effective aggregator in this study. It combines GCN Aggregator~\cite{kipf2016semi} and GraphSage Aggregator~\cite{hamilton2017inductive} by:

\begin{equation} 
    \label{eq:biInteraction}
    \begin{split}
    f_\text{Bi-Interaction} = & \; \text{LeakyReLU}(\mathbf{W}_{\text{agg(1)}}(\mathbf{h}^{*} + \mathbf{h}_{Ego})) \; + \\
    & \; \text{LeakyReLU}(\mathbf{W}_{\text{agg(2)}}(\mathbf{h}^{*} \odot \mathbf{h}_{Ego})),
    \end{split}
\end{equation}

\where LeakyReLU is used as an activation function. Also, $\mathbf{W}_{\text{agg(1)}}, \mathbf{W}_{\text{agg(2)}} \in \mathbb{R}^{m\times m}$ are trainable weight matrices.

\subsection{Recommendation-Making Layer}

For each of the higher-propagation layers, the embeddings of $u$ and $i$ are obtained as follows: 

\begin{equation} 
    \label{eq:embeddingConcat}
    \textbf{u}^* = \textbf{u}^{*(0)} \Vert \cdots \Vert \textbf{u}^{*(L)}, \;\; \textbf{i}^* = \textbf{i}^{*(0)} \Vert \cdots \Vert \textbf{i}^{*(L)},
\end{equation}
\where $\textbf{u}^*$ and $\textbf{i}^*$ are the concatenation of fused user and item embeddings from each layer with totally $L$ layers. Then, the final prediction score of a user-item interaction is computed by:

\begin{equation} 
    \label{eq:bi_interaction}
    \hat{y}(u,i) = {\textbf{u}^*}^\top \cdot \textbf{i}^*.
\end{equation}

Moreover, the objection function of the recommendation-making layer with Bayesian Personalized Ranking (BPR)~\cite{rendle2012bpr} loss is defined as:

\begin{equation} 
    \label{loss:Pred}
    \mathcal{L}_{\text{Pred}} = \sum_{(u,i,j)\in{\Omega}} -\ln \,\sigma\, \left( \hat{y}(u,i) - \hat{y}(u,j) \right),
\end{equation}

\where $\Omega = {(u,i,j)|(u,i) \in \mathcal{T^+}, (u,j) \in \mathcal{T^-})}$ denotes training set. Here, $\mathcal{T^+}$ indicates the observed interactions between user $u$ and item $i$. $\mathcal{T^-}$ indicates the sampled unobserved interactions between user $u$ and item $j$. Finally, the overall model objective function is constructed as:

\begin{equation} 
    \label{loss:CF}
    \mathcal{L}_{\text{KGIF}} = \mathcal{L}_{\text{Embed}} + \mathcal{L}_{\text{Pred}} + \lambda \,\Vert \Theta \Vert^{2}_{2},
\end{equation}
\where $\Theta = \{\mathbf{E}, \mathbf{h}_{p}, \mathbf{t}_{p}, \mathbf{W}_1, \mathbf{W}_2, \mathbf{W}_{\text{agg(1)}}^{(l)}, \mathbf{W}_{\text{agg(2)}}^{(l)} \mid \forall l \in \{1, \dots, L\} \}$ is the set of parameters, and $\mathbf{E}$ is the CKG embeddings. $L_2$ regularization with $\lambda$ is applied to prevent overfitting.

\section{Experiments}\label{Experiments}

\begin{table}[t]
    \centering
    \caption{Statistics of the experimental datasets}
    \label{tab:default}
    \begin{tabularx}{\columnwidth}{cXccc}
    \toprule
    & & Amazon-book & Last-FM & Yelp2018 \\
    \midrule \midrule
    \multirow{4}{*}{\shortstack{User-Item \\ Interaction}}  
    & \#Users & 70,679 & 23,566 & 45,919 \\ 
    & \#Items & 24,915 & 48,123 & 45,538 \\
    & \#Interactions & 847,733 & 3,034,796 & 1,185,068 \\
    & Density & 4.81e-04 & 2.68e-03 & 5.67e-04 \\
    \midrule
    \multirow{4}{*}{\shortstack{Item \\ Attributes}} 
    & \#Entities & 88,572 & 58,266 & 90,961 \\
    & \#Triplets & 2,557,746 & 464,567 & 1,853,704 \\ 
    & \#Relations & 39 & 9 & 42 \\ 
    & Density & 2.97e-05 & 1.84e-05 & 1.07e-05 \\
    \midrule
    CKG & Density & 3.88e-05 & 1.33e-04 & 1.73e-05 \\
    \bottomrule
    \end{tabularx}
    \label{Data_Stats}
\end{table}

We evaluate the performance of KGIF using three widely recognized benchmark data sets: Amazon-book, Last-FM, and Yelp2018. In addition, the proposed model is studied with two types of experiments: 1) comparative experiments with eight benchmark models for overall performance and 2) ablation experiments for analyzing the effectiveness of each component in KGIF. In the ablation study, we investigate: 
\begin{itemize}
    \item \textbf{AB-01}: Effectiveness of projection vectors.
    \item \textbf{AB-02}: Impacts of the explicit side-information fusion layer
    \item \textbf{AB-03}: Performance analysis of ego-network orders.
\end{itemize}
This section contains the following parts: 1) details of data sets, 2) evaluation metrics, 3) benchmark model introduction, 4) comparative experiments and analysis, 5) ablation study, and 6) a visual interpretation of the recommendation-making with a case study.

\subsection{Data and Experimental Setup}

\begin{table*}[t]
    \begin{threeparttable}
    \centering
    \caption{Overall performance comparison between KGIF and all baseline models}
    \label{table:PerformanceComparison}
    \begin{tabularx}{\textwidth}{ll|cccccccccc|A}
    \toprule 
    Dataset & Metric & FM & NFM & RippleNet & GC-MC & KGNN-LS & KGAT & CKAN & KGCL & KGIN & KGRec & \textbf{KGIF} \\
    \midrule \midrule
    \multirow{2}{*}{Amazon-book} 
    & Recall & 0.1345 & 0.1366  & 0.1336 & 0.1316 & 0.1362 & 0.1489 & 0.1497 & 0.1569 & \underline{0.1613} & 0.1609 & \textbf{0.1614}\\
    & NDCG & 0.0886 & 0.0913 & 0.0910 & 0.0874 & 0.0860 & 0.1006 & 0.1014 & 0.1093 & 0.1101 & \underline{0.1106} & \textbf{0.1112}\\
    \midrule
    \multirow{2}{*}{Last-FM} 
    & Recall & 0.0778 & 0.0829 & 0.0791 & 0.0818 & 0.0880 & 0.0870 & 0.0882 & 0.0907 & 0.0900 & \underline{0.0943} & \textbf{0.0946}\\
    & NDCG & 0.1181 & 0.1214 & 0.1238 & 0.1253 & 0.1253 & 0.1242 & 0.1290 & 0.1353 & \underline{0.1395} & 0.1392 & \textbf{0.1424}\\
    \midrule
    \multirow{2}{*}{Yelp2018} 
    & Recall & 0.0627 & 0.0660 & 0.0664 & 0.0659 & 0.0637 & 0.0693 & 0.0668  & \underline{0.0695} & 0.0680 & 0.0682 & \textbf{0.0697}\\
    & NDCG & 0.0768 & 0.0810 & 0.0822 & 0.0790 & 0.0802 & 0.0844 & 0.0802  & 0.0813 & \underline{0.0845} & 0.0842 & \textbf{0.0852}\\
    \bottomrule
    \end{tabularx}
    \begin{tablenotes}
    \small
    \item The highest-performing entries are highlighted in Bold, while the second-highest performances are indicated with an Underline.
    \end{tablenotes}
    \end{threeparttable}
\end{table*}

We conduct experiments on three widely used benchmark data sets: \textbf{Amazon-book}~\cite{he2016ups}, a subset selected from Amazon-review~\cite{he2016ups}, for product recommendation research; \textbf{Last-FM}~\cite{Celma:Springer2010}, a subset of the original data set, containing the data from January 2015 to June 2015; \textbf{Yelp2018}~\cite{wang2019neural}, a subset from the 2018 Yelp Challenge edition. 

\tabref{Data_Stats} presents basic dataset statistics. We analyze the data densities across three categories: 1) user-item (BG), 2) item-attribute (KG), and 3) overall (CKG). The analysis reveals that 1) Amazon-book has the highest item-attribute density and volume, 2) Last-FM exhibits the highest user-item and overall densities, and 3) Yelp2018 shows the lowest overall and item-attribute densities.

To ensure consistency, we preprocess each dataset using the N-core setting, where only users and items with at least N interactions are retained. We set $N=10$, ensuring that only users and items with 10 or more interactions are included. This preprocessing is applied uniformly to all models to maintain relevance and reduce data sparsity.

For each dataset, we randomly split the user-item interactions into three parts with a $7:2:1$ ratio: 70\% for training, 20\% for testing, and 10\% for validation to tune hyperparameters. Positive samples are defined by observed user-item interactions, and for each, we generate a negative sample by pairing the user with an item they have not interacted with.

\subsection{Evaluation Metrics}

In our evaluation, we adopt the standard approach of computing preference scores for all items for each user, excluding interactions present in the training set. For each user in the test set, the model predicts scores for items that were not seen during training, thereby generating a ranked list of potential recommendations. The effectiveness of top-K recommendations and rankings is measured using Recall@$K$ and NDCG@$K$, with $K=20$ chosen as the default to strike a balance between capturing relevant recommendations and maintaining computational efficiency. NDCG is particularly critical for assessing the accuracy of the rankings, as it accounts not only for the relevance of recommended items but also for their order, making it an essential metric in scenarios where the sequence of recommendations significantly influences user satisfaction and decision-making. 

\subsection{Baseline Models}

We assess the efficacy of KGIF by comparing it with various existing baseline models based on different technical perspectives. These include factorization-based models like FM (Factorization Model) and NFM (Neural Factorization Model), regularization and path-based approaches, and methods that take advantage of GNNs and GATs. Specifically, we consider:

\noindent \textbf{Supervised Learning methods:}
\begin{itemize}[leftmargin=*]    
    \item \textbf{FM}~\cite{rendle2011fast} and \textbf{NFM}~\cite{he2017neural} use fundamental factorization techniques.
\end{itemize}

\noindent \textbf{Path-based Recommenders:}
\begin{itemize}[leftmargin=*]    
    \item \textbf{RippleNet}~\cite{wang2018ripplenet} integrates aspects of both regularization and path-based methodologies.
\end{itemize}

\noindent \textbf{GNN-based Recommenders:}
\begin{itemize}[leftmargin=*]    
    \item \textbf{GC-MC}~\cite{van2018graph} deploys a GCN encoder to interpret the user-item interaction graph.
    \item \textbf{KGNN-LS}~\cite{wang2019knowledge} is a GNN-based model, that transforms a KG into user-specific graphs. It takes into account user preferences on KG relations and label smoothness during the information aggregation phase to produce user-specific item representations.
    \item \textbf{KGAT}~\cite{wang2019kgat} uses GAT to integrate the side information during the information propagation process implicitly.
    \item \textbf{CKAN}~\cite{wang2020ckan} employs distinct neighborhood aggregation schemes on the user-item graph and the KG to derive user and item embeddings.
    \item \textbf{KGIN}~\cite{wang2021learning} models user-item interactions at a fine-grained level by using intents as attentive combinations of KG relations. It incorporates a new GNN-based information aggregation scheme that integrates relation sequences of long-range connectivity, improving both model capability and interpretability. 
\end{itemize}

\begin{figure*}[t]
    \centering
    \includegraphics[width=\textwidth]{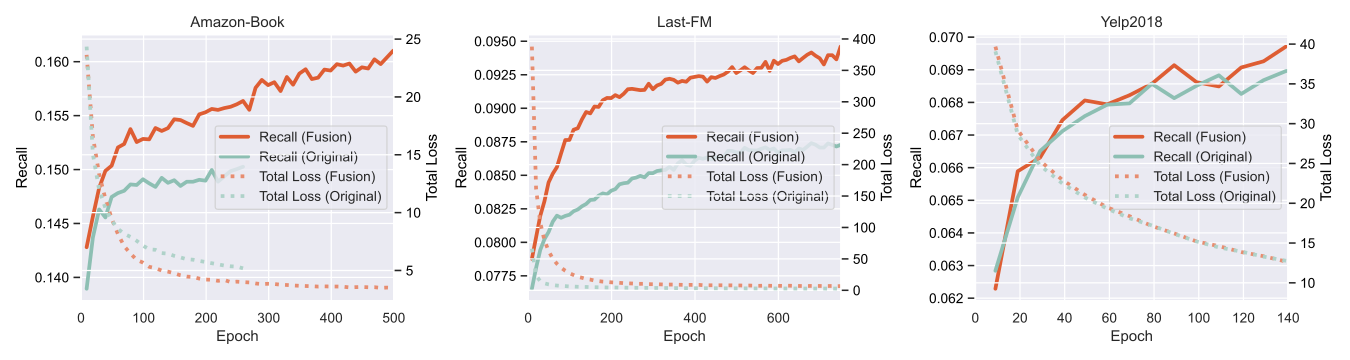}
    \caption{Comparative Analysis of Loss and Recall: Fusion vs. Without Fusion (Original) Approach}
    \label{fig:recall_loss}
\end{figure*}
    
\noindent \textbf{Self-Supervised Recommenders:}
\begin{itemize}[leftmargin=*]    
    \item \textbf{KGCL}~\cite{yang2022knowledge} leverages contrastive learning (CL) as auxiliary information, integrating KG learning with user-item interaction modeling within a unified self-supervised learning framework.
    \item \textbf{KGRec}~\cite{yang2023knowledge} introduces a self-supervised rationalization method for recommender systems, using an attentive mechanism to score and utilize key knowledge triplets in generative and contrastive tasks, thereby enhancing both the knowledge graph and the user-item graph.
\end{itemize}

\subsection{Hyperparameter Settings}

All models were configured with an embedding size of 64, initialized using the Xavier method~\cite{glorot2010understanding}. Optimization was performed with the Adam optimizer~\cite{kingma2014adam}, using a learning rate of 0.0001 and a batch size of 1024. For baseline models, the original hyperparameter settings were retained to ensure optimal performance as reported in their respective papers. Pre-trained MF embeddings were used to enhance stability and accelerate convergence. Through grid search, $L_2$ regularization was tuned to a value of $10^{-4}$ to prevent overfitting, and a dropout ratio of 0.1 was applied at each layer. Early stopping was employed, halting training if the recall@20 on the validation set showed no improvement for 50 consecutive epochs. 

In KGIF, we modeled fourth-order connectivity by setting the depth $L$ to four, with hidden dimensions of 64, 32, 16, and 8. The impact of layer depth on performance is discussed further in Section 4.8.

\tabref{table:reproduce} lists the hyperparameters for KGIF across the three datasets, including learning rate $\rho$, node dropout ratio $p_{\text{drop}}$, embedding size $d$, number of aggregation layers $L$, and $L_2$ regularization coefficients $\lambda^{\text{user}}_2$ and $\lambda^{\text{item}}_2$ for user and item embeddings.

\begin{table}[t]
    \centering
    \caption{Hyperparameter settings of KGIF}
    \label{table:reproduce}
    \begin{tabularx}{\columnwidth}{lXXXXXXX}
    \toprule &
    $\rho$ & $p_{\text{drop}}$ & $d$ & $L$ & $\lambda^{\text{user}}_2$ & $\lambda^{item}_2$ \\    
    \midrule \midrule
    Amazon-book & $10^{-4}$ & 0.1 & 64 & 4 & $10^{-5}$ & $10^{-5}$ \\
    Last-FM & $10^{-4}$ & 0.1 & 64 & 4 & $10^{-5}$ & $10^{-5}$ \\
    Yelp2018 & $10^{-4}$ & 0.1 & 64 & 4 & $10^{-4}$ & $10^{-4}$ \\
    \bottomrule
    \end{tabularx}
\end{table}

\subsection{Comparative Experiments}

The comprehensive performance results are shown in \tabref{table:PerformanceComparison}, demonstrating that KGIF consistently outperforms all baseline models across the three benchmark datasets in terms of Recall@20 and NDCG@20. Three key observations stand out:

\begin{itemize}[leftmargin=*] \item KGIF delivers the best performance across all datasets. This success validates the proposed CKG embedding with explicit information fusion, which captures intricate entity interactions and efficiently leverages collaborative signals. Unlike traditional attention mechanisms, KGIF employs compositional projection vectors and explicit fusion for better initial CKG embedding and improved use of item-attribution information. For example, KGIF excels on the Amazon-book dataset, which has the richest item-attribution information. This allows KGIF to fully exploit the detailed attribute information and achieve significant gains in both Recall and NDCG. Even on the Last-FM dataset—where item-attribute interactions are sparse but user-item interactions are dense—KGIF still outperforms other models. On Yelp2018, the most sparse dataset, KGIF achieves top results in both Recall and NDCG, demonstrating robustness in sparse environments.

\item Self-supervised models like KGCL and KGRec also perform well, owing to their use of contrastive learning, which effectively filters out long-tailed item attributes and noisy user-item interactions. This method improves model robustness, particularly in sparse environments such as Yelp2018. However, KGIF consistently surpasses these models, suggesting that explicit information fusion further enhances the model’s ability to manage noisy data and extract meaningful patterns. KGIF's ability to dynamically adapt to both user-item and item-attribute relationships gives it an edge over self-supervised methods, especially in datasets like Amazon-book, where the volume and diversity of item attributes are high.

\item KGIF's superior performance over GNN-based models such as KGAT and CKAN highlights the importance of initial CKG embedding. By utilizing dynamic projection vectors, KGIF can capture complex relational patterns more effectively, leading to better higher-order information aggregation. In Last-FM, where user-item interactions are dense but item-attribute information is sparse, KGIF's dynamic projection vectors allow it to maintain its performance advantage. The explicit information fusion in KGIF ensures that the model can incorporate both direct and indirect connections, thereby improving the quality of its recommendations compared to the static embeddings used by other GNN-based models. \end{itemize}

\subsection{AB-01 Effectiveness of Projection Vectors} 
\label{sec:projectionVectors}

\begin{figure}[t]
    \centering
    \includegraphics[width=1.0\columnwidth]{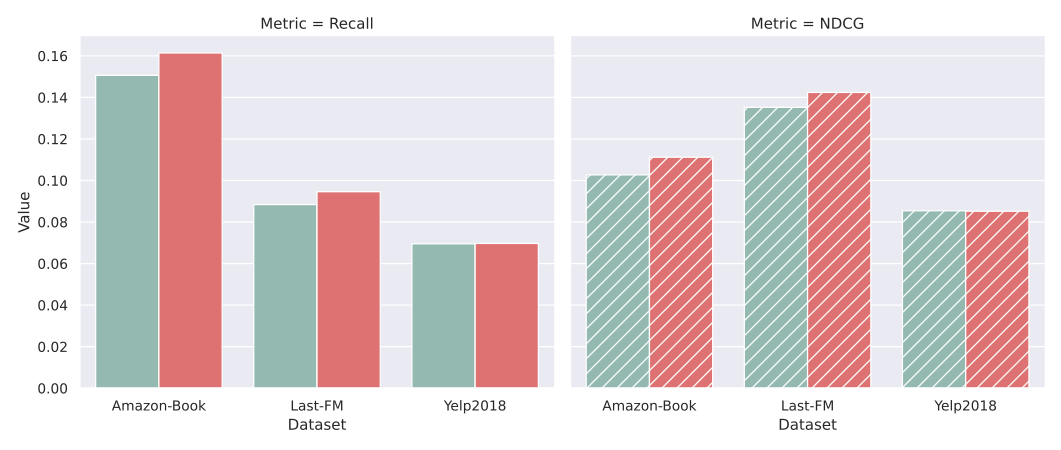}
    \caption{\small{Performance of Fusion with Various Embedding Methods.}}
    \label{fig:performance_metric}
\end{figure}

This section explores the effectiveness of using projection vectors over projection matrices for CKG embedding. Dynamic projection vectors offer a more precise and flexible representation of the CKG by adapting to specific relational contexts, unlike the static transformation matrices used in methods like TransR. To validate this, two classical KG embedding methods are compared: TransD, which utilizes projection vectors, and TransR, which employs transformation matrices. As illustrated in \figref{fig:performance_metric}, embedding fusion combined with TransD consistently outperforms TransR across all three datasets. This improvement is driven by the dynamic nature of projection vectors, which tailor the embeddings for both head and tail entities based on their specific relationships, enabling more context-aware embedding fusion. This approach is particularly effective for handling complex relation types such as $1$-to-$1$, $1$-to-$J$, $J$-to-$1$, and $J$-to-$J$. KGIF, when integrated with TransD and explicit fusion, consistently demonstrates superior performance, highlighting the role of dynamic projection vectors in capturing diverse and intricate relational patterns within the knowledge graph.

\begin{table}[t]
    \begin{threeparttable}
        \centering
        \caption{Effect of different fusion methods}
        \label{table:fusionMethods}
        \begin{tabularx}{\columnwidth}{XcXXXXXX}
            \toprule 
            \multicolumn{1}{l}{\multirow{2}{*}{Fusion type}} &
            \multicolumn{1}{c}{\multirow{2}{*}{SW}} &
            \multicolumn{2}{c}{\small{Amazon-book}} &
            \multicolumn{2}{c}{\small{Last-FM}} &
            \multicolumn{2}{c}{\small{Yelp2018}} \\
            \cmidrule{3-4} \cmidrule{5-6} \cmidrule{7-8}
            \multicolumn{1}{l}{} & & Recall & NDCG & Recall & NDCG & Recall & NDCG \\
            \midrule \midrule
            \multicolumn{2}{l}{Without fusion} & 0.1540 & 0.1046 & 0.0924 & 0.1398 & 0.0692 & 0.0835 \\
            \midrule
            \multirow{2}{*}{Addition} 
                & $\ocircle$ & 0.1580 & 0.1065 & 0.0933 & 0.1421 & 0.0693 & 0.0840 \\
                & $\times$ & \underline{0.1599} & \underline{0.1095} & 0.0940 & \underline{0.1426} & \underline{0.0694} & \underline{0.0849} \\
            \midrule
            \multirow{2}{*}{Concatenation} 
                & $\ocircle$ & 0.1582 & 0.1077 & 0.0922 & 0.1403 & 0.0690 & 0.0836 \\
                & $\times$ & 0.1583 & 0.1081 & 0.0929 & 0.1414 & 0.0690 & 0.0838 \\
            \midrule
            \multirow{2}{*}{Multiplication} 
                & $\ocircle$ & 0.1584 & 0.1076 & \underline{0.0944} & \textbf{0.1434} & 0.0692 & 0.0839 \\
                & $\times$ & \textbf{0.1614} & \textbf{0.1112} & \textbf{0.0946} & 0.1424 & \textbf{0.0697} & \textbf{0.0852} \\
            \bottomrule
        \end{tabularx}
        \begin{tablenotes}
        \small
        \item SW: Shared weights
        \end{tablenotes}
    \end{threeparttable}
\end{table}

\subsection{AB-02 Impacts of Information Fusion Layer} \label{sec:embeddingFusion}

This section provides detailed evidence of the effectiveness of the information fusion layer, focusing on how dynamic projection vectors facilitate reparameterization during the fusion process. The novel fusion layer introduced in KGIF explicitly incorporates side information, which prior methods implicitly handled, leading to suboptimal utilization of relational data.

As shown in \tabref{table:fusionMethods}, various fusion strategies were evaluated, including additive and multiplicative methods, both with and without shared weights. Our results demonstrate that multiplicative fusion without shared weights consistently yields the best performance across all datasets (Amazon-book, Last-FM, and Yelp2018). This approach achieved the highest Recall and NDCG scores across the board. For example, on Amazon-book, the Recall improved from 0.1540 (without fusion) to 0.1614, and NDCG increased from 0.1046 to 0.1112, validating the superiority of this method.

The graph analysis (\figref{fig:recall_loss}) highlights how the fusion-based KGIF model demonstrates faster loss reduction and higher Recall compared to the original non-fusion models. On Amazon-book, the fusion model not only stabilizes at a higher Recall but also shows a significantly faster decrease in Total Loss, indicative of better optimization. Similarly, for Last-FM, the fusion-based model achieves both a higher and more stable Recall, along with a sharper reduction in Total Loss. Although Yelp2018 shows less dramatic gains, the improvements in Recall are still consistent, and the loss declines more rapidly during the early training stages.

\subsection{AB-03 Performance Analysis of Ego-network Orders}

Our performance analysis provided in \tabref{table:EgoLayers} reveals that KGIF benefits from increased network depth up to 4 layers, beyond which baseline models~\cite{wang2019kgat, wang2020ckan, yang2022knowledge} typically experience diminishing returns and performance degradation. This advantage is due to KGIF's ability to propagate information effectively through explicit information fusion and dynamic projection vectors. As the network deepens, KGIF continues to capture valuable insights from both direct and indirect relationships, leading to consistent improvements in Recall and NDCG. Unlike other models that suffer from oversmoothing at similar depths, KGIF maintains the relevance of its embeddings by integrating side information and relational data at each layer. The use of dynamic projection vectors preserves the heterogeneity of node features, adapting to relational contexts within the knowledge graph and enabling KGIF to utilize a 4-layer architecture without performance degradation.

\begin{table}[t]
    \centering
    \caption{Effect of embedding propagation layer numbers}
    \label{table:EgoLayers}
    \begin{tabularx}{\columnwidth}{lXXXXXX}
    \toprule &
    \multicolumn{2}{c}{Amazon-book} &
    \multicolumn{2}{c}{Last-FM} &
    \multicolumn{2}{c}{Yelp2018} \\
    \cmidrule(lr){2-3}\cmidrule(lr){4-5}\cmidrule(lr){6-7} &
    Recall & NDCG & Recall & NDCG & Recall & NDCG \\
    \midrule \midrule
    1-Layer & 0.1468 & 0.0993 & 0.0909 & 0.1375 & 0.0689 & 0.0835 \\
    2-Layer & 0.1567 & 0.1072 & 0.0926 & 0.1404 & 0.0690 & 0.0840 \\
    3-Layer & 0.1607 & 0.1100 & 0.0936 & 0.1414 & 0.0693 & 0.0841 \\
    4-Layer & \textbf{0.1614} & \textbf{0.1112} & \textbf{0.0946} & \textbf{0.1424} & 0.0697 & \textbf{0.0852} \\
    5-Layer & 0.1608 & 0.1103 & 0.0938 & 0.1418 & \textbf{0.0698} & 0.0843 \\
    \bottomrule
    \end{tabularx}
\end{table}

\subsection{Recommendation-Making Interpretation}

\begin{figure}[t]
    \centering
    \includegraphics[width=0.8\columnwidth]{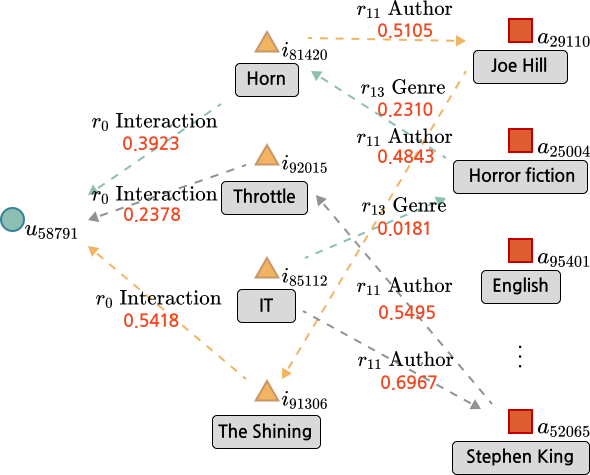}
    \caption{\small{Visual Interpretation of Recommendation-Making with Attention Scores}}
    \label{fig:attention_score}
\end{figure}

The high complexity of the deep learning model has introduced great difficulty in model explanation~\cite{xian2019reinforcement, ma2019jointly}. While KGIF is highly complex in this study, we visually interpret the recommendation-making process with attention scores in a case study for user $u_{58791}$ in \figref{fig:attention_score}. The figure shows how the novel \textit{IT} by \textit{Stephen King} is recommended based on the cumulative attention scores along various relational paths. For instance, the relationship between \textit{IT} and \textit{Stephen King} as the author yields a significant attention score of 0.6967, while additional contributions come from the interaction with \textit{The Throttle} (0.2378) and another author-related score from \textit{The Shining} (0.5495). These scores are aggregated, resulting in a strong cumulative score that justifies recommending \textit{IT} to the user.

\section{Conclusion and future work}\label{Conclusion}

In this study, we introduced KGIF, a novel framework for knowledge-aware recommender systems, designed to efficiently manage the complex web of interactions within CKGs. By employing explicit information fusion and a reparameterization technique through dynamic projection vectors, KGIF effectively captures both direct and indirect relational patterns, enhancing the model's ability to handle heterogeneous data. Our experimental results demonstrate that KGIF significantly improves recommendation accuracy, achieving superior Recall and NDCG scores compared to traditional models, while adeptly navigating the intricate relationships present in CKGs.

Moving forward, while over-smoothing~\cite{chen2020measuring}—the phenomenon where node features become overly similar with increasing network depth—did not substantially impact our current findings, particularly in the densely connected Yelp2018 dataset, it remains an area for further investigation. Future work will explore techniques to mitigate over-smoothing, such as introducing novel graph convolutional strategies or applying depth-wise regularization. These enhancements will further strengthen the ability of recommendation systems to process and extract meaningful insights from increasingly complex knowledge graphs.
\section*{Acknowledgment}
\noindent This work was supported by the Department of Computer Science at Bowling Green State University.

\newpage

\bibliographystyle{IEEEtran}
\bibliography{References}
\end{document}